\documentclass[letterpaper]{article} 
\usepackage{aaai2026}  
\usepackage{times}  
\usepackage{helvet}  
\usepackage{courier}  
\usepackage[hyphens]{url}  
\usepackage{graphicx} 
\usepackage{natbib}  
\usepackage{caption} 
\usepackage{algorithm}
\usepackage{algorithmic}

\usepackage{newfloat}
\usepackage{listings}

\usepackage{float} 
\usepackage{array}
\usepackage[utf8]{inputenc}
\usepackage[T1]{fontenc} 
\usepackage{multicol} 
\usepackage{enumitem}   
\usepackage{subcaption}
\usepackage{booktabs} 
\usepackage{tabularx} 
\usepackage{multirow}
\usepackage{longtable}
\usepackage{ragged2e}
\usepackage[table,xcdraw]{xcolor}
\usepackage{longtable}
\usepackage{booktabs}
\usepackage{array}

\usepackage[edges]{forest}
\usepackage{pdflscape}

\newcolumntype{C}{>{\centering\arraybackslash}m{2.2em}} 

\usepackage{amssymb}

\newcommand{\rothead}[1]{\makebox[9pt][l]{\rotatebox{40}{#1}}}

\newcommand{\covbox}[2]{\raisebox{-1pt}{{\setlength{\fboxsep}{0pt}\setlength{\fboxrule}{0.6pt}\fcolorbox{#1}{#2}{\rule{0pt}{6.4pt}\rule{6.4pt}{0pt}}}}}
\newcommand{\full}{\covbox{black}{black}}
\newcommand{\partm}{\covbox{black!35}{black!35}}
\newcommand{\nomatch}{\covbox{black!50}{white}}

\DeclareCaptionStyle{ruled}{labelfont=normalfont,labelsep=colon,strut=off} 
\floatstyle{ruled}
\newfloat{listing}{tb}{lst}{}
\floatname{listing}{Listing}
\title{Unaccountable Delegation, Fading Skills: \\Mapping the Risks of Workplace AI Agents}

\author {
    Gabriele La Malfa\textsuperscript{\rm 1},
    Lakmal Meegahapola	\textsuperscript{\rm 2},
    Edyta Bogucka\textsuperscript{\rm 2},
    Jie M. Zhang\textsuperscript{\rm 1},
    Michael	Luck\textsuperscript{\rm 3},
    Elizabeth Black\textsuperscript{\rm 1},
    Daniele	Quercia\textsuperscript{\rm 2, 4}
}
\affiliations {
    \textsuperscript{\rm 1}King's College London, UK\\
    \textsuperscript{\rm 2}Nokia Bell Labs, Cambridge, UK\\
    \textsuperscript{\rm 3}University of Sussex, UK\\
    \textsuperscript{\rm 4}Politecnico di Torino, Italy\\
    \{gabriele.la\_malfa, jie.zhang, elizabeth.black\}@kcl.ac.uk, michael.luck@sussex.ac.uk, \\\{lakmal.meegahapola, edyta.bogucka, daniele.quercia\}@nokia-bell-labs.com
}

\usepackage{bibentry}
\nocopyright
\begin{document}

\maketitle


\vspace{-0.2 in}
\begin{abstract}

To anticipate the socio-technical risks posed by AI agents, organizations first need taxonomies to classify them. However, existing AI risk taxonomies focus on broad risks and do not capture the job-specific risks introduced by agents. To address this gap, we make three main contributions. First, we developed a multi-layer framework based on a review of the literature on AI agents. The framework models three core components and their interactions: the agents, their goals, and their environment. Second, we embedded this framework in a structured prompt and applied it to descriptions of 2,078 job tasks from the O*NET occupational database, producing 8,356 risk scenarios labeled by severity and deployment mode (automation or augmentation). We validated these scenarios with 45 workers across 10 job roles and an independent LLM judge, confirming their high plausibility and alignment with the corresponding job tasks. Finally, we extended an existing taxonomy to create a 15-category taxonomy of workplace AI agent risks that covers all our risk scenarios. Our analysis highlights four findings. First, augmentation is not inherently safe because overreliance on agents can gradually erode workers' skills and oversight. Second, erroneous actions of agents account for the largest share of risk scenarios and has the highest number of severe risks. Many of these risks arise at the human--agent boundary. Third, automation is associated mainly with organizational risks (operational failures, financial losses), whereas augmentation is associated mainly with risks to workers (capability erosion, psychological and social risks). Fourth, workers found our taxonomy easier to use for a risk classification task than two other risk taxonomies and preferred it in 64\% of non-tied comparisons with a recent generative AI risk taxonomy. Taken together, these findings show that workplace AI agent risks do not arise from agents alone; they also depend on how people work with agents and how agents are deployed. Safer workplaces therefore require not only safer agents but also carefully designed human--AI agent collaboration. {Supplementary material and further details available at:} 
\mbox{\url{https://social-dynamics.net/ai-risks/workplace-agents}}

\end{abstract}

\section{Introduction}

\emph{AI agents} are autonomous entities that perceive their environment, interact with humans or other agents, and act to achieve goals with varying degrees of independence~\cite{Sapkota_2026}. They are increasingly deployed to support everyday workplace tasks~\cite{eloundou2024gpts, xi2025rise, mohney2025what}. AI agents do not operate in isolation: together with the environment they perceive, the goals they pursue, and the humans they work alongside, they form what we call an \emph{agentic AI system} (AAIS)~\cite{IBMAgenticAI}. As these systems are deployed across jobs and industries, organizations need ways to identify and classify the risks they introduce~\cite{weidinger2023sociotechnicalsafetyevaluationgenerative}. This need arises when teams red-team agents before launch~\cite{ganguli2022red}, conduct impact assessments~\cite{moss2021assembling}, or conduct risk and incident audits~\cite{raji2020closing, mcgregor2021preventing}. In all of these cases, organizations rely on risk taxonomies: classification systems that provide a shared language for anticipating risks that have not yet occurred and classifying those that have~\cite{shelby2023sociotechnicalharmsalgorithmicsystems}. Without a taxonomy built for the specific context of workplace AI agents, these efforts rely on risk classifications that can leave risks unnamed and untracked.

Existing AI risk taxonomies were not designed for workplace agentic AI systems, leaving three gaps. First, many taxonomies are too generic or focus primarily on technical failures. They can therefore overlook \emph{socio-technical AI risks} that emerge from the dynamic interaction between an AI agent and its social context, including the humans, organizations, and institutions in which it is embedded. Second, most taxonomies were developed for generic AI~\cite{abercrombie2024collaborativehumancentredtaxonomyai} or generative AI~\cite{li2025closerlookexistingrisks}, rather than for AI agents, which can act more autonomously. Even recent AI agent risk taxonomies tend to model agents as purely technical systems~\cite{hammond2025multiagentrisksadvancedai, MicrosoftTaxonomy2025, khoo2025greatcapabilitiescomegreat, ghosh2025safetysecurityframeworkrealworld}, without explicitly capturing how agents, goals, environments, and human workers interact to produce risks. Third, taxonomies based on incident databases and media reports~\cite{abercrombie2024collaborativehumancentredtaxonomyai, turri2023we, paeth2025lessons} capture risks that have already occurred and were visible enough to be recorded. They are therefore less suited to identifying risks that develop gradually or have not yet produced visible harms, such as the erosion of professional skills caused by prolonged reliance on AI~\cite{richards2025incidents, Yiduo_Shao2025}. Together, these gaps point to the need for a taxonomy that operates at the level of specific job tasks, reflects how agentic AI systems are structured, and anticipates risks before they become harms. To address these gaps, we make three main contributions:

\noindent\textbf{1. We introduce a multi-layer framework for workplace AI agents.}
Building on prior work on socio-technical safety and compositional risk assessment, the framework organizes risks across three layers (\S\ref{The Multi-Layer Risk Analysis Framework}): {Technical Capability} (AAIS components and how they work together), {Human Interaction} (how people engage with AAIS), and {Systemic Impact} (broader organizational and societal consequences). The framework also captures the relationships among agents, goals, environments, and human workers. We refined the framework through structured interviews with three experts in agentic AI systems and responsible AI.

\vspace{0.05 in}
\noindent\textbf{2. We generate and validate job-specific risk scenarios.}
We embedded the framework in a structured prompt and applied it to descriptions of 2,078 job tasks from the O*NET occupational database (\S\ref{subsec:corpus}). This process generated 8,356 risk scenarios labeled by severity (minimal, limited, high, critical) and deployment mode (\S\ref{subsec:generation}): augmentation, in which an agent supports a worker, or automation, in which the agent performs the job task without worker involvement. We then validated the scenarios in two ways (\S\ref{subsec:risk-validation}). First, 45 workers across 10 job roles evaluated 450 scenarios from their own roles. Second, an LLM-as-a-judge from a different model family assessed the full corpus. Both evaluations found high plausibility (workers: 4.03/5; LLM: 4.81/5) and strong task alignment (workers: 3.94/5; LLM: 4.60/5).

\vspace{0.05 in}
\noindent\textbf{3. We construct and validate a workplace AI agent risk taxonomy.}
To ground the taxonomy in both generated and observed risks, we supplemented the 8,356 generated scenarios with 222 workplace AI agent incidents from three public databases and 11 failure cases from a live multi-agent red-teaming study. We converted these cases into a common risk scenario format and used the combined corpus to extend Li et al.'s generative AI risk taxonomy into a workplace AI agent risk taxonomy (\S\ref{subsec:taxonomy-construction}). The resulting taxonomy contains 15 categories and 44 sub-categories (Figure~\ref{Workplace-Taxonomy}). We evaluate its structural qualities, compare its coverage with 10 established AI risk frameworks, and test its usability in a comparative study with 26 workers (\S\ref{subsec:taxonomy-validation}). Finally, we analyze the risk corpus along the themes of taxonomy categories, severity, deployment mode, and framework layers.

\vspace{0.05 in}

Across these contributions, our results point to a clear pattern (\S\ref{sec:results}): most workplace AI agent risks arise from how agents and people work together. The two largest taxonomy categories are \emph{Erroneous Actions} of agents (30.6\%), including misinterpretations and incorrect recommendations, and \emph{Capability Erosion} (21.3\%), including the loss of workers' skills and oversight capabilities. Together, these categories account for 51.9\% of all risk scenarios and occur mostly under augmentation. Under automation, risks shift to the organizational level, particularly \emph{Operational Failures} and \emph{Financial Losses}. Taken together, these findings suggest that safer workplaces require not only safer agents but also carefully designed human--AI agent collaboration.

\section{Related Work}\label{sec:related_work} 
Prior work provides a good foundation for understanding AI risks in general. However, three gaps remain. Existing taxonomies tend to operate at the wrong level of detail for specific job tasks (\S\ref{subsec:gap1}), are rarely built around how agents behave (\S\ref{subsec:gap2}), and focus on past incidents rather than emerging risks (\S\ref{subsec:gap3}).

\subsection{From Broad Risk Classifications to Job-Level Risk}\label{subsec:gap1}

Research on AI risk tends to fall into two areas: technical and socio-technical. Both offer valuable perspectives, but neither maps easily onto what happens when a specific AI agent enters a specific job workflow. Technical approaches explain risks primarily at model-level, such as bias in training data or failures of robustness~\cite{10.1145/3457607, Raji_2022, 10.1145/3551624.3555286}. These are important concerns, but they tend to abstract away the organizational and human context in which AI operates. Socio-technical approaches take a broader view, framing AI as part of a wider social system and classifying large-scale risks such as unfair resource distribution or the erosion of accountability~\cite{10.1145/3287560.3287598, https://doi.org/10.1111/risa.13850, shelby2023sociotechnicalharmsalgorithmicsystems, slattery2025airiskrepositorycomprehensive}. These approaches show why context matters, but their broad categories can be difficult to apply to a particular task or workflow. Neither perspective is designed to answer a practical question facing organizations: ``What risks could emerge when this AI agent is introduced into this specific job task?'' What is missing is a framework that connects multi-layer risk concepts to concrete job tasks and workflows.

\subsection{Modeling Risks from Agent Behavior}\label{subsec:gap2}

Risks of AAISs can emerge from how agents pursue goals, take multi-step actions, and interact with their environments~\cite{Sapkota_2026, Chan_2023, gridach2025agenticaiscientificdiscovery, liu2025advanceschallengesfoundationagents}. When several agents work together within an AAIS, coordination failures, conflicting goals, and emergent behavior create additional risks~\cite{hammond2025multiagentrisksadvancedai, altmann2024emergencemultiagentsystemssafety, kong2025surveyllmdrivenaiagent}. Autonomy can also make agent behavior harder to monitor and responsibility harder to assign~\cite{anwar2024foundationalchallengesassuringalignment, ranjan2025fairnessagenticaiunified}. Moreover, recent work categorizes agent-specific risks such as misalignment with human intent, prompt-injection vulnerabilities, and cascading failures in multi-agent systems~\cite{MicrosoftTaxonomy2025, khoo2025greatcapabilitiescomegreat, ghosh2025safetysecurityframeworkrealworld}. However, these risk classifications often study agents in isolation or in simulated environments. They rarely connect the three components at the core of an AAIS: the agents, their goals, and their environment, to concrete job tasks performed by workers. Thus, existing agent-safety research identifies important risks and failure mechanisms but does not yet explain how those mechanisms produce risks in specific job tasks.

\subsection{From Existing Evidence to Anticipating Risks}
\label{subsec:gap3}

AI risk taxonomies draw on several sources of evidence. Documentation-based approaches capture reported incidents and other public records~\cite{li2025closerlookexistingrisks, lee2024deepfakesphrenologysurveillancemore}, academic and policy literature~\cite{shelby2023sociotechnicalharmsalgorithmicsystems, slattery2025airiskrepositorycomprehensive}, or user-generated data such as social media discussions~\cite{zhang2025darkside}. Human-centered approaches develop or refine risk categories through interviews, expert consultation, workshops, and annotation studies~\cite{steenstra2025riskontology, abercrombie2024collaborativehumancentredtaxonomyai}. Some studies combine these sources. For example, Abercrombie et al.~\cite{abercrombie2024collaborativehumancentredtaxonomyai} build on existing taxonomies and documented incidents, then refine their categories through expert feedback and crowdsourced annotation. These sources offer complementary benefits. Documented evidence grounds taxonomies in observed harms, while human elicitation can reveal contextual knowledge that public records overlook. However, both also have limitations. Incident databases contain only harms that were observed, reported, and made public~\cite{paeth2025lessons, richards2025incidents}. Literature-based taxonomies inherit the scope of prior research, while human elicitation depends on the experience and composition of the participants. As a result, emerging, gradual, and job-specific risks may remain underrepresented, particularly as AI capabilities and workplace uses evolve. Prospective methods offer a complementary way to identify risks before they produce documented harms. Research in labor economics and HCI has used the O*NET occupational database to ground analyses of AI exposure in real job tasks~\cite{eloundou2024gpts, felten2023will}. Other work shows that LLMs can support structured scenario generation and red-teaming exercises that anticipate risks before deployment~\cite{perez2022red, ganguli2022red, frohling2026agent, 10.5555/3716662.3716711}. However, these research strands have largely developed separately. What is missing is a systematic method that combines real job-task data with prospective risk generation to identify workplace AI agent risks before they become documented harms.

Taken together, prior work offers broad risk classifications, detailed accounts of agent failure modes, and strong methods for learning from existing evidence. However, these three strands have not yet been fully integrated. A workplace AI agent risk taxonomy must therefore (1) be developed by considering specific job tasks, (2) consider risks across \emph{Technical Capability}: AAISs including agents, goals, and environments, and how they interact; \emph{Human Interaction}: how workers engage with these systems; and \emph{Systemic Impact}: broader organizational and societal consequences, and (3) anticipate risks that may not yet appear in incident records. Together, these gaps motivate our job-task-grounded and forward-looking approach, which combines anticipated risk scenarios with documented evidence to construct a taxonomy of workplace AI agent risks.
\section{Methodology}\label{sec:methodology}

\begin{figure*}[t]
\centering
\includegraphics[width=0.62\textwidth]{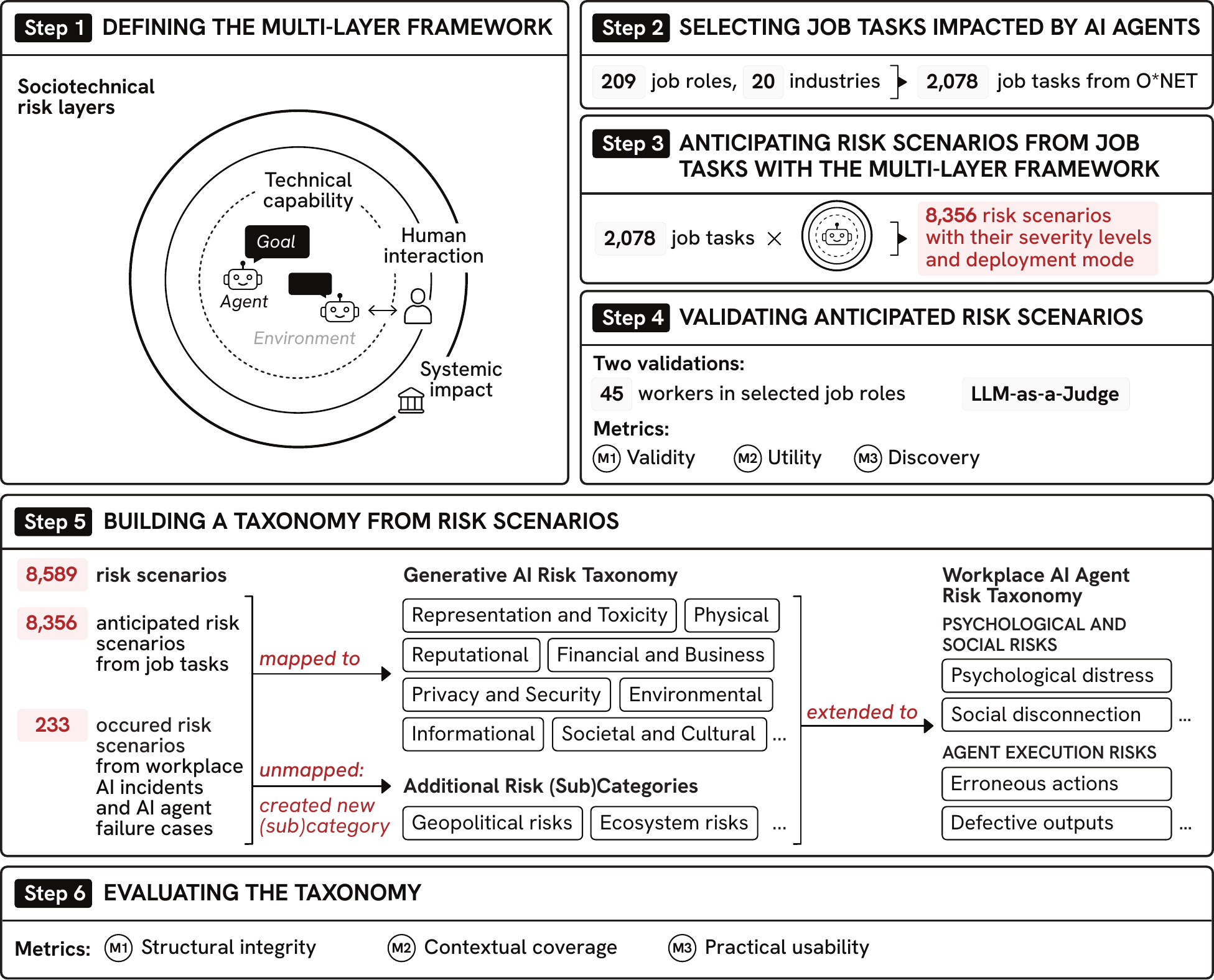}
\caption{\textbf{Overview of the six-step methodology, from framework definition through risk scenario generation and validation to taxonomy construction and evaluation.}
\textbf{Step 1} defines the multi-layer framework, which organizes sociotechnical risk into three layers: \emph{technical capability} (AAIS including the AI agents, goals they pursue, and the environment they act in), \emph{human interaction} (how workers engage with the AAIS and components), and \emph{systemic impact} (consequences that propagate beyond the individual job task).
\textbf{Step 2} selects 2,078 computer-based O*NET job tasks (209 roles, 20 industries).
\textbf{Step 3} applies the framework to each task through a structured prompt, anticipating 8,356 risk scenarios labeled with severity and deployment mode.
\textbf{Step 4} validates the scenarios with 45 workers from the selected roles and with an LLM-as-a-judge, on validity, utility, and discovery metrics.
\textbf{Step 5} builds the taxonomy from 8,589 scenarios: the 8,356 anticipated ones plus 233 describing risks that have already occurred, drawn from workplace AI incidents and documented agent failure cases; scenarios that do not map onto the GenAI Risk Taxonomy produce new categories and sub-categories, extending it into the 15-category Workplace AI Agent Risk Taxonomy.
\textbf{Step 6} evaluates the taxonomy on structural integrity, contextual coverage, and practical usability.}
\label{Prompt-Baseline}
\vspace{-10pt}
\end{figure*}

Our methodology consists of six steps (Figure~\ref{Prompt-Baseline}). First, we define a multi-layer framework for analyzing risks in AAISs (\S\ref{The Multi-Layer Risk Analysis Framework}). Second, we select job tasks in the O*NET database that AI agents are likely to affect (\S\ref{subsec:corpus}). Third, we embed the multi-layer framework in a structured prompt to generate risk scenarios for job tasks (\S\ref{subsec:generation}). Fourth, we evaluate the scenarios with workers and an LLM-as-a-judge (\S\ref{subsec:risk-validation}). Fifth, we use the risk scenarios to construct a workplace AI agent risk taxonomy (\S\ref{subsec:taxonomy-construction}). Finally, we evaluate the taxonomy for structural integrity, coverage relative to 10 established frameworks, and practical usability (\S\ref{subsec:taxonomy-validation}).

\subsection{Defining the Multi-Layer AAIS Framework}
\label{The Multi-Layer Risk Analysis Framework}

We build the multi-layer framework by extending the sociotechnical framework of Weidinger et al.~\cite{weidinger2023sociotechnicalsafetyevaluationgenerative}, which argues that evaluating technical artifacts alone cannot establish whether an AI system is safe. A capability may indicate a potential hazard, but whether it produces harm depends on the people using the system, the context of use, and the broader systems in which it is deployed. The framework therefore distinguishes three targets of evaluation: \emph{capability}, which examines the AI system and its technical artifacts; \emph{human interaction}, which examines how people use and respond to the system; and \emph{systemic impact}, which examines broader organizational, social, economic, and environmental effects. These layers add progressively more context, but they are not chronological stages and can influence one another. To apply this layered approach to workplace AI agents, we make the internal structure of an agentic AI system explicit. Workplace agents pursue goals, act in changing environments, and interact with other agents and human workers. Risks can arise from these components or from their interactions, even when each component appears to function correctly. We therefore retain the three evaluation layers while expanding them to represent the components and interaction pathways of an AAIS.

\vspace{0.05in}
\noindent\textbf{Layer 1: Capability.}
This layer represents risks arising from three components and their interactions~\cite{10.5555/1671238, 10.5555/1695886}: the \emph{AI agents}, their \emph{environment}, and their \emph{goals}.

\emph{Component risks.} Agent properties such as autonomy, reactivity, and proactiveness can create different risks. An agent may respond poorly to an unfamiliar situation, take an unrequested action, or modify a workflow without sufficient oversight. The environment, including APIs, databases, and physical workspaces~\cite{IBMAgenticAI}, can create risks when it provides incomplete, outdated, or rapidly changing information. Goals can create risks when they are represented by poor proxy measures. For example, a diagnostic agent optimized for case volume may deprioritize difficult cases to increase throughput.

\emph{Interaction risks.} Risks can also emerge from interactions among components. In \emph{agent--agent} interactions, missing coordination rules may lead agents to take conflicting actions. In \emph{agent--environment} interactions, an agent may optimize a local objective while violating environmental or organizational constraints. In \emph{agent--goal} interactions, an agent may satisfy an objective in an unintended way. For example, a quality-control agent could classify every item as defective to maximize its detection score, thereby halting production.

\vspace{0.05in}
\noindent\textbf{Layer 2: Human Interaction.}
This layer represents how workers interact with agents and the broader system~\cite{hammond2025multiagentrisksadvancedai}. We distinguish five interaction pathways. In \emph{agent--human} interactions, agents may undermine trust, violate workplace norms, or encourage over-reliance. \emph{Human--human} interactions may change when AI-generated targets influence how managers evaluate and direct workers. \emph{Human--environment} interactions capture how agent-oriented workplaces may constrain or endanger workers, such as when a warehouse is organized primarily for robotic efficiency. \emph{Human--goal} interactions capture conflicts between system objectives and professional judgment, such as when clinical staff feel pressured to follow automated triage scores. Finally, \emph{human--AAIS} interactions capture risks arising from the system as a whole, including cases in which responsibility becomes too distributed for workers to understand or contest a decision.

\vspace{0.05in}
\noindent\textbf{Layer 3: Systemic Impact.}
This layer represents consequences that extend beyond an individual task or interaction. Such risks may affect organizations, industries, society, the economy, or the natural environment. They often emerge through large-scale adoption, repeated interactions, or feedback loops rather than a single failure. For example, interacting AI trading agents may collectively amplify incorrect market signals and contribute to a market disruption. Widespread use of AI management systems may shift bargaining power toward employers or normalize harmful workplace practices. Similarly, repeated use of agents that favor established solutions may gradually reduce the diversity of professional knowledge across a field.

\vspace{0.05in}
\noindent\textbf{Framework Review and Refinement.}
We reviewed the framework through three structured interviews with experts in agentic AI systems and responsible AI. Each interview lasted approximately 30 minutes and included a framework briefing, technical questions, and detailed feedback (Supplementary Material B). The experts assessed the coherence of the components, interaction pathways, and three layers and identified areas requiring clarification. We used their feedback to refine definitions, clarify the boundaries between interactions, and improve the presentation of the framework.

\subsection{Selecting Job Tasks Impacted by AI Agents}\label{subsec:corpus}

To ground our analysis in real job tasks, we used O*NET~\cite{onet_database_2025}, a U.S. government database describing the tasks, activities, and work contexts of hundreds of jobs. We used a filtered subset developed by Shao et al.~\cite{shao2025futureworkaiagents}, which focuses on computer-based tasks on which AI agents are likely to be deployed. It contains 2,078 job tasks from 209 job roles across 20 industries, including healthcare, finance, and transportation. Each task is described concretely. Examples include ``Diagnose acute conditions that could result in rapid physiological deterioration'' for an acute care nurse and ``Examine whether the organization's objectives are reflected in its management activities'' for a manager. These descriptions provide the job context needed to generate job task-specific risk scenarios.

\subsection{Generating Risk Scenarios for Job Tasks with the Multi-Layer Framework}\label{subsec:generation}

We use {gpt-4o-mini}~\cite{gpt-mini} to generate prospective 
risk scenarios for each of the 2,078 tasks. This approach is 
consistent with prior work showing that LLMs can serve as effective 
tools for structured risk foresight~\cite{bogucka2024atlas, 
buçinca2023ahafacilitatingaiimpact, 10.5555/3716662.3716711, 
Wang_2024_Farsight, perez2022red, frohling2026agent}. 
We selected {gpt-4o-mini} to balance reasoning quality 
with the computational demands of processing thousands of prompts 
at scale. To prevent the model from producing generic or 
implausible outputs, we used a structured prompt that included a description of our multi-layer framework defining agentic AI systems' risk surfaces, enabling the generation of risks specific to agents and their interactions (Supplementary Material E; Figure~\ref{Prompt-Baseline}). The 
prompt instructed the model to consider risks at each 
framework layer (capability, human interaction, systemic impact) 
and to label each scenario with its corresponding component or 
interaction pathway (e.g., agent--human, agent--goal). This structure kept the generated risks tied to specific pathways rather than generic failures. For each task, the model was asked to generate up 
to five risk scenarios, enough to capture a diverse set of 
risks while avoiding highly speculative edge cases. Each 
scenario was annotated with two additional labels: (1) 
\textit{severity}, classified according to four risk tiers (minimal, limited, high, critical)~\cite{yin2025bingoguard}, 
and (2) \textit{deployment mode}, specifying whether the risk 
arises from \textit{automation} (complete replacement of the 
human task as defined in ~\cite{raisch2021artificial}) or \textit{augmentation} (AI supporting the human in 
performing the tasks as defined in ~\cite{autor2015there}). This process produced 8,356 distinct risk scenarios.

\subsection{Validating Risk Scenarios}\label{subsec:risk-validation}

We assessed the quality of the generated scenarios using both workers and an LLM judge, in line with previous work~\cite{10.5555/3666122.3668142, li2024llmsasjudgescomprehensivesurveyllmbased, chiang2023largelanguagemodelsalternative, li2025generationjudgmentopportunitieschallenges, frohling2026agent}. In both cases, each scenario was evaluated on 10 metrics across three dimensions: \emph{validity}, covering plausibility and connection to the task; \emph{utility}, covering usefulness, actionability, detail, complexity, and specificity; and \emph{discovery}, covering originality, rarity, and novelty~\cite{si2024llmsgeneratenovelresearch, hu2024unveilingllmevaluationfocused}.

\vspace{0.03 in} \noindent\textbf{Domain Expert Evaluation.} We selected 10 job roles and recruited 45 workers whose reported professions matched those roles (Supplementary Material C). We collected demographic and role-related information from each expert. For each job role, we randomly sampled 10 of the approximately 70 role-specific scenarios. Each expert independently evaluated the 10 scenarios for their role using the 10 metrics described above on a 5-point Likert scale. In total, the workers evaluated 450 risk scenarios.

\vspace{0.03 in}

\vspace{0.03 in}
\noindent\textbf{LLM-as-a-Judge Evaluation.}
The domain expert evaluation provided useful ratings but covered only the subset of scenarios that the experts could feasibly review. To rate all 8,356 scenarios using the same criteria, we used {gemini-3-flash}~\cite{gemini} as an LLM judge. The model rated each scenario using the 10 metrics described above, allowing us to apply a consistent standard across the full corpus. We took two steps to assess the reliability of these ratings. First, we chose a judge from a different model family than the generator (Google Gemini rather than the OpenAI model used to generate the scenarios), since models tend to score their own outputs more favorably~\cite{10.5555/3737916.3740113}. Second, we tested the judge's consistency by re-rating a random sample of 370 scenarios with a second model, {gpt-4o}, and comparing the two. The models rarely gave the exact same 5-point score. However, when we collapsed the 5-point scale into positive, neutral, and negative bands (\emph{sentiment agreement}), the two models landed in the same band almost every time: 98.9\% for plausibility and 98.1\% for connection to the task. Therefore, the judges agreed on the underlying judgment for the validation sample (Supplementary Material C).

\begin{figure*}[ht]
    \centering
    \includegraphics[width=0.8\textwidth]{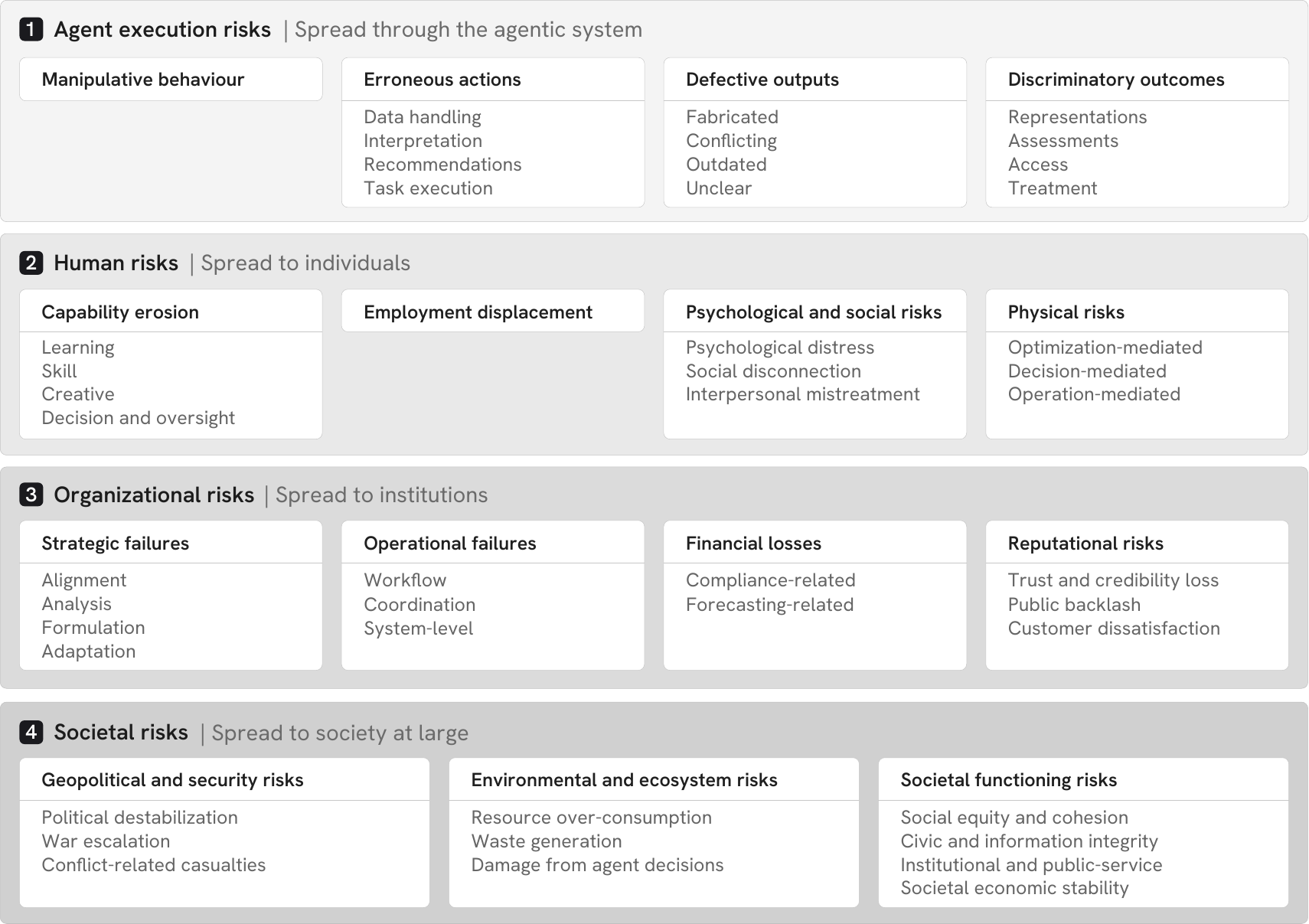}
    \caption{\textbf{Workplace AI Agent Risk Taxonomy.} 15 
    categories and 44 sub-categories organized across agent execution risks, human risks, organizational risks, and societal risks.}
    \label{Workplace-Taxonomy}
\end{figure*}

\vspace{0.03 in}
\vspace{0.03 in} \noindent\textbf{Results.} Both the workers and 
the LLM judge rated the risk scenarios as highly 
plausible (workers: 4.03/5; LLM: 4.81/5) and closely tied to the 
corresponding job tasks (workers: 3.94/5; LLM: 4.60/5). They also 
rated the scenarios as useful (workers: 3.70/5; LLM: 3.38/5) and actionable (workers: 3.62/5; LLM: 3.37/5). Their ratings differed most on novelty. workers considered the scenarios moderately novel (originality: 2.96/5; rarity: 3.00/5), while the LLM judged them as more familiar (originality: 1.80/5; rarity: 1.28/5). Overall, we interpret these results as a sign that the risk scenarios are well grounded: workers recognized them as plausible and practically relevant, while the LLM, which has seen vast amounts of text, recognized them as patterns it had encountered before (Supplementary Material C).


\subsection{Building a Taxonomy from Risk Scenarios}
\label{subsec:taxonomy-construction}

We began with the 8,356 risk scenarios generated in the previous step. To ground the taxonomy in real-world evidence, we added 11 AI agent failure cases from a study of agent behavior~\cite{shapira2026agentschaos} and 222 workplace AI incidents from three public incident databases. We converted both the failure cases and incidents into the same risk scenario format as the generated scenarios. Next, our goal was to map as many scenarios as possible to an existing taxonomy. We used the Generative AI (GenAI) risk taxonomy of Li et al.~\cite{li2025closerlookexistingrisks} as the starting point and followed three steps. First, we used {gpt-4o-mini} (for the generated risk corpus) or manual inspection (for failure cases and incidents) to map each scenario to the closest category or sub-category in Li et al.'s taxonomy. Scenarios without a suitable match were marked as unmatched. Second, we analyzed the unmatched scenarios and identified recurring themes. Third, we manually reviewed these themes to determine how they should extend the taxonomy. When a theme fit within an existing category, we added it as a new sub-category. When it did not fit any existing category, we created a new top-level category. We then refined the category and sub-category definitions to make their boundaries clear. The resulting Workplace AI Agent Risk Taxonomy contains 15 top-level categories and 44 sub-categories (Figure~\ref{Workplace-Taxonomy}). See Supplementary Material G for the full taxonomy with descriptions and example risk scenarios.

\subsection{Evaluating the Taxonomy}\label{subsec:taxonomy-validation}

We evaluated the taxonomy in three ways. First, we checked whether the categories were distinct (structural integrity). Second, we checked whether the taxonomy captured workplace risk categories that existing frameworks overlook (contextual coverage). Third, we examined whether workers could use it to classify workplace risks (practical usability).

\vspace{0.03 in} \noindent\textbf{Structural Integrity.} We first examined whether the 15 categories and 44 sub-categories were distinct in both their vocabulary and meaning. \emph{Vocabulary check.} We represented the taxonomy with names and definitions for categories and sub-categories. Then, we converted this text into a TF--IDF vector, which gives more weight to words that are distinctive to a category and less weight to commonly used words. We then calculated the cosine similarity between every pair of categories on a scale from 0 to 1, where higher values indicate greater overlap. The mean inter-category similarity was 0.01, indicating minimal overlap in vocabulary. We also calculated Jaccard similarity, which measures the proportion of words shared by two descriptions. Using 0.30 as the overlap threshold, 99.9\% of sub-category pairs fell below this value. \emph{Meaning check.} Categories may use different words while describing similar concepts. For example, ``skill erosion'' and ``deskilling'' have similar meanings despite sharing few words. To capture this type of overlap, we encoded the category definitions using the pre-trained sentence-transformer model: {all-MiniLM-L6-v2}. The model represents texts with similar meanings as nearby vectors, even when they use different words. The mean pairwise semantic similarity was 0.24, indicating limited average overlap in meaning. Together, these checks suggest that the taxonomy definitions are distinct in both their vocabulary and their underlying concepts (Supplementary Material H).

\begin{table*}[h]
\centering
\footnotesize
\setlength{\tabcolsep}{2.2pt}
\setlength{\extrarowheight}{2.6pt}
\begin{tabular}{l *{10}{c}}
    \textbf{Workplace AI Agent Risk Category} &
    \rothead{EPIC} & \rothead{CSET} & \rothead{Turing} & \rothead{Azure} & \rothead{Sony} & \rothead{TASRA} & \rothead{Shelby} & \rothead{AIR} & \rothead{MA Adv AI} & \rothead{AI Risk Repo}\\
    \midrule
    \multicolumn{11}{@{}l}{\textbf{Agent Execution Risks}} \\
    \quad Manipulative Behaviour & \partm & \nomatch & \partm & \partm & \partm & \partm & \partm & \partm & \full & \full \\
    \quad Erroneous Actions & \nomatch & \nomatch & \partm & \nomatch & \partm & \partm & \partm & \full & \partm & \full \\
    \quad Defective Outputs & \nomatch & \nomatch & \nomatch & \nomatch & \nomatch & \partm & \full & \nomatch & \partm & \full \\
    \quad Discriminatory Outcomes & \full & \full & \full & \full & \full & \full & \full & \full & \partm & \full \\
    \addlinespace[3pt]
    \multicolumn{11}{@{}l}{\textbf{Human Risks}} \\
    \quad Capability Erosion & \partm & \nomatch & \partm & \full & \partm & \partm & \full & \partm & \nomatch & \full \\
    \quad Employment Displacement & \full & \nomatch & \nomatch & \full & \full & \partm & \partm & \partm & \nomatch & \full \\
    \quad Psychological and Social Risks & \full & \full & \full & \full & \full & \partm & \full & \nomatch & \partm & \nomatch \\
    \quad Physical Risks & \full & \full & \full & \full & \full & \partm & \full & \full & \partm & \partm \\
    \addlinespace[3pt]
    \multicolumn{11}{@{}l}{\textbf{Organizational Risks}} \\
    \quad Strategic Failures & \nomatch & \nomatch & \nomatch & \partm & \nomatch & \nomatch & \partm & \nomatch & \partm & \nomatch \\
    \quad Operational Failures & \nomatch & \partm & \nomatch & \nomatch & \nomatch & \nomatch & \nomatch & \nomatch & \partm & \nomatch \\
    \quad Financial Losses & \full & \full & \nomatch & \full & \nomatch & \full & \nomatch & \partm & \partm & \nomatch \\
    \quad Reputational Risks & \full & \full & \full & \nomatch & \full & \full & \partm & \nomatch & \partm & \nomatch \\
    \addlinespace[3pt]
    \multicolumn{11}{@{}l}{\textbf{Societal Risks}} \\
    \quad Geopolitical and Security Risks & \nomatch & \nomatch & \nomatch & \nomatch & \nomatch & \full & \nomatch & \full & \partm & \nomatch \\
    \quad Environmental and Ecosystem Risks & \partm & \full & \full & \full & \nomatch & \full & \full & \nomatch & \partm & \full \\
    \quad Societal Functioning Risks & \full & \partm & \partm & \full & \full & \full & \full & \full & \partm & \full \\
    \bottomrule
\end{tabular}
\caption{\textbf{Coverage of our 15 taxonomy categories across
ten established AI risk classifications.} Categories are grouped by
dimension. We use the color coding: \full~= full match, \partm~=
partial match, \nomatch~= no match. Compared risk classifications
are: EPIC~\cite{Fergussonetal2023}, CSET~\cite{HoffmanFrase2023}, the
Alan Turing Institute~\cite{https://doi.org/10.5281/zenodo.3240529},
Microsoft Azure~\cite{Microsoft2023}, Sony~\cite{Hutiri_2024},
TASRA~\cite{critch2023tasrataxonomyanalysissocietalscale}, and Shelby
et al.~\cite{shelby2023sociotechnicalharmsalgorithmicsystems}, AIR
2024~\cite{zeng2024airiskcategorizationdecoded}, Hammond et
al.~\cite{hammond2025multiagentrisksadvancedai}, and the AI Risk
Repository~\cite{slattery2025airiskrepositorycomprehensive}. We find
that organizational risks (\emph{Strategic Failures},
\emph{Operational Failures}) and agent-execution risks
(\emph{Erroneous Actions}, \emph{Defective Outputs}) are
consistently absent from or only partially covered by prior work.}
\label{tab:taxonomy_contextualization}
\end{table*}

\vspace{0.03 in}
\noindent\textbf{Contextual coverage.}
To check whether our taxonomy captures workplace risk categories that prior work overlooks, we compared it with ten established AI risk taxonomies and classification frameworks, which we collectively call \emph{risk classifications}. We followed the comparative mapping approach of Abercrombie et al.~\cite{abercrombie2024collaborativehumancentredtaxonomyai}. Our comparison included the seven sources examined in their study and we also included three more recent sources (Table~\ref{tab:taxonomy_contextualization}). \emph{Mapping process.} For each of our 15 categories, we examined its coverage in each of the ten risk classifications. We assigned one of three labels: \emph{Full match} (\full) when a comparison source contained a category that directly represented ours; \emph{Partial match} (\partm) when it covered only part of our category; and \emph{No match} (\nomatch) when it did not cover the risk (Supplementary Material L). Table~\ref{tab:taxonomy_contextualization} shows that prior risk classifications cover human-oriented risks thoroughly: \emph{Discriminatory Outcomes} of agents is fully represented in nine of the ten classifications, \emph{Physical Risks} in seven, and \emph{Psychological and Social Risks} in six. Beyond these, three gaps emerge. First, organizational risks are the least covered in the comparison: \emph{Operational Failures} and \emph{Strategic Failures} are the only two categories with no full match anywhere, and are absent from eight and seven of the ten classifications respectively. Second, the agent-execution risks that dominate our corpus are recognized only sporadically: under agent execution risks, \emph{Defective Outputs} is absent from six classifications, and \emph{Erroneous Actions} is fully represented in only two, appearing elsewhere as generic model inaccuracy rather than as distinct failures of what an agent interprets, recommends, and produces. Third, and most tellingly, \emph{Manipulative Behaviour} of agents is fully represented only in the two classifications built specifically for agentic systems~\cite{hammond2025multiagentrisksadvancedai, slattery2025airiskrepositorycomprehensive}; the remaining eight treat deception either as human misuse of a model or not at all. By contrast, our \emph{Societal Functioning Risks} category is well covered by prior work, with seven full matches, which reinforces that its small share in our corpus (0.4\%) reflects our task-level elicitation rather than a gap in the literature. Together, these patterns suggest that existing risk classifications were not designed to represent the organizational, operational, and agent-execution risks that emerge when AI agents are introduced into workplaces. Our workplace-specific taxonomy addresses this limitation.

\vspace{0.03 in} \noindent\textbf{Practical usability.} A taxonomy should be easy to use in practice. We therefore designed a study around a common risk-management activity: classifying risks with a taxonomy. This task is routine in impact assessments, model cards, or risk registers~\cite{mitchell2019modelcardsmodelreporting, gebru2018datasheetsdatasets, moss2021assembling, raji2020closing}. Using the same professional-background matching criteria as in the scenario-validation study, we recruited 26 workers from five professional backgrounds: software engineering, data science, design, healthcare, and finance. All participants confirmed daily AI use and medium-to-high awareness of AI agents and passed three comprehension checks on the definition of AI agents before proceeding. Each participant classified five AI agent risk scenarios drawn at random from a held-out pool not used during taxonomy construction. The pool combined LLM-generated scenarios with documented incidents from public AI incident databases, so that participants saw both anticipated risks and real failures. Each scenario included its workplace context: the risk, industry, job title, and corresponding job task. Participants classified every scenario using three taxonomies: (A) our \emph{Workplace AI Agent Risk Taxonomy} with 15 categories and 44 sub-categories; (B) the \emph{Generative AI Risk Taxonomy}~\cite{li2025closerlookexistingrisks} with 12 categories and 48 sub-categories; and (C) the \emph{AI Risk Repository Domain Taxonomy}~\cite{slattery2025airiskrepositorycomprehensive} with 7 categories and 23 sub-categories. We presented the taxonomies in a randomized order under the neutral labels Framework 1, Framework 2, and Framework 3 to reduce name recognition bias. For each scenario and taxonomy, participants selected the best fitting category or chose ``no good fit'' when no category represented the risk. This option allowed us to distinguish between difficulty selecting among relevant categories and the absence of a suitable category. Participants then rated the ease of using each taxonomy on a seven-point scale and provided a brief written reflection on their experience using different taxonomies for risk classification. We report the findings in \S\ref{sec:results}.

\begin{figure*}[tbp]
    \centering
    \includegraphics[width=\linewidth]{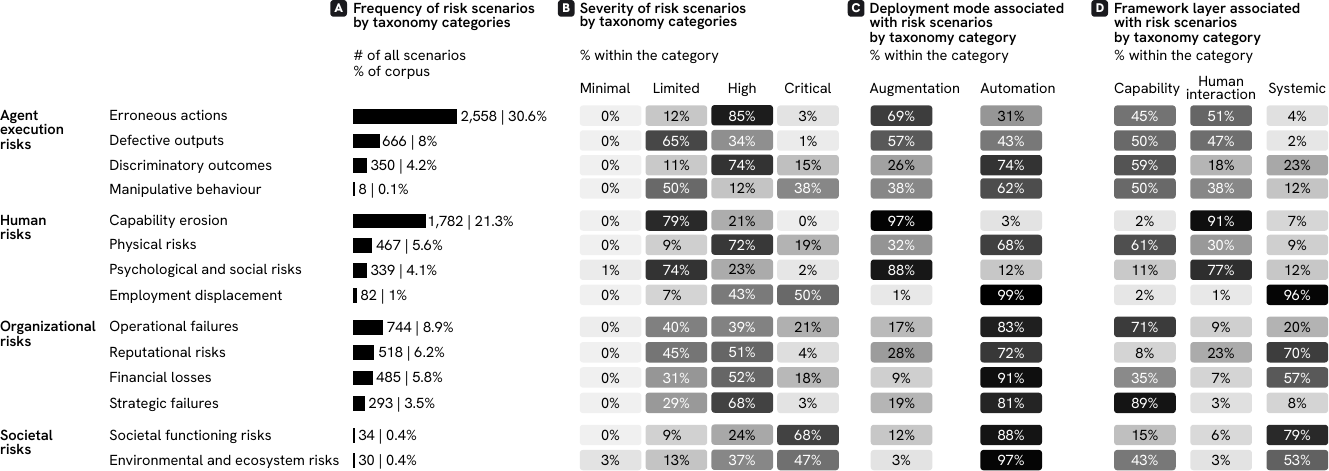}
    \caption{\textbf{Distribution of 8,356 workplace AI agent risk scenarios across the taxonomy categories.}
\textbf{(A)} Number of scenarios in each category, with its share of the full corpus; categories are ordered by frequency, under 4 themes.
\textbf{(B)} Severity composition of each category.
\textbf{(C)} Deployment mode: whether the agent assists a worker (augmentation) or replaces one (automation).
\textbf{(D)} Framework layer at which the risk arises: technical capability, human interaction, or systemic impact.
In panels (B)--(D), cells are row percentages and shading is proportional to the value.
\emph{Geopolitical and Security Risks} is omitted because no generated scenario was assigned to it; this category and its sub-categories were formed from incident-based risk scenarios rather than generated ones.}
    \label{FrequencyandHetmaps}
\end{figure*}
\section{Main Findings}\label{sec:results}

We classified our 8,356 risk scenarios into the
taxonomy and examined how they distribute across taxonomy categories,
severity levels, deployment modes, and framework layers (Figure~\ref{FrequencyandHetmaps}). Our main findings are provided below.

\paragraph{Finding 1: Augmentation is not inherently safe; it 
introduces a slow and largely invisible risk of deskilling.}

AI augmentation, where the agent assists rather than replaces the 
human, is often treated as the more responsible deployment 
choice~\cite{wang2024investigating, shao2025futureworkaiagents}. 
Our findings complicate this assumption. \emph{Capability
Erosion} is the second most common risk category in the entire
dataset, representing 21.3\% (1,782) of all scenarios, and 97.0\% of
these risks arise from augmentation rather than
automation, the most one-sided augmentation profile in the taxonomy.
Within it, the dominant sub-category is not the loss of a manual
skill but the loss of oversight itself: \emph{Decision and oversight} in \emph{Capability Erosion} alone accounts for 1,420 scenarios (17.0\% of the corpus), describing workers who stop questioning, checking, or
independently judging what the agent produces. The pattern is consistent with earlier research on human factors, which has long warned that the 
more reliable a system is, the more people stop practicing the 
skills they would need if it ever failed~\cite{bainbridge1983ironies}. What makes modern AI agents different is that they are proactive rather than passive: they suggest next steps, flag priorities, and take initiative, which accelerates the rate at which workers hand over their professional judgment to the system. The severity profile confirms that this is not a dramatic or sudden failure: 79\% of capability erosion risks are classified as ``limited'' severity, and only $\le$0.1\% as
``critical''. The risk is real precisely because it is gradual,
cumulative, and easy to miss until it is too late to reverse.

\paragraph{Finding 2: Erroneous agent action is both the most common
and the most severe risk, and it is not purely a technical problem.}

\emph{Erroneous Actions} under \emph{Agent Execution Risks}, which covers situations where an
agent mishandles data, misreads a situation, advises incorrectly, or
executes a task wrongly, is the largest category in the taxonomy at
30.6\% (2,558) of all scenarios, and it also carries the highest
number of severe risks: 88\% are rated ``high'' or
``critical'' (2,183 ``high''). This might seem like a
straightforward technical issue: the model hallucinates, or its
training data is wrong. But our analysis shows that the failure
happens between the agent and the human, not inside the model
itself. Fabrication is rare, accounting for only 57 scenarios
(0.7\% of the corpus, filed under \emph{Defective Outputs} in \emph{Agent Execution Risks}). The dominant sub-types are instead \emph{erroneous interpretation}, where the
agent misreads data, context, or intent (1,158 scenarios, 13.9\%),
and \emph{erroneous recommendations}, where it advises incorrectly
on that basis (946 scenarios, 11.3\%). Moreover, 69.5\% of these
risks arise under augmentation, meaning a human is present and
acting on the flawed output. The agent produces an answer that is technically consistent with its inputs but wrong for the situation; the human acts on it because 
the agent communicated confidence. Improving model accuracy alone
will therefore not solve the problem.

\paragraph{Finding 3: Automation concentrates risk in the 
organization; augmentation concentrates it in the worker, and 
each requires different governance.}

The capability erosion and agent error risks above are concentrated
in different deployment modes, and a third pattern only becomes
visible when we consider automation. When agents fully
automate a task, removing humans from the loop, the
dominant risks are organizational and institutional: \emph{Financial
Losses} (485 risks, 90.9\% from automation), \emph{Operational
Failures} (744 risks, 83.5\%), and \emph{Reputational Risks} (518
risks, 71.8\%). The starkest case is \emph{Employment Displacement},
which is 98.8\% automation-bound and 92.7\% high or critical
severity, the most one-sided and most severe category in the
taxonomy. These failures tend to be visible, fast, and
expensive: legal liability, operational downtime, or public
relations damage from an agent acting without oversight. When agents augment rather than automate, the risks land on the 
individual worker rather than the organization, manifesting as the 
slow erosion of professional agency described in Finding 1. This
structural divide has a direct governance implication: a single
approach cannot cover both deployment modes; we translate this into
concrete safeguards in the Discussion.

\paragraph{Finding 4: The Workplace AI Agent Risk Taxonomy 
outperforms existing frameworks in practical usability.}

In our usability study, the Workplace AI Agent Risk Taxonomy was the
clearest of the three taxonomies for classifying workplace AI agent
risks. Across 26 workers classifying 130 risks, the Workplace AI Agent Risk Taxonomy 
achieved a higher ease-of-use rating than the GenAI Risk 
Taxonomy~\cite{li2025closerlookexistingrisks} (mean 5.75 vs. 5.46 
on a 1--7 scale; parametric: $t(129) = 2.23$, $p = .027$, Cohen's 
$d = 0.20$; non-parametric: Wilcoxon $W = 1300.5$, $p = .026$). A 
Friedman test across all three frameworks was also significant 
($\chi^2(2) = 6.33$, $p = .042$). Our taxonomy was preferred in 
64\% of decided pairwise comparisons against the GenAI Risk 
Taxonomy (54 wins vs. 30, 46 ties) and achieved the highest 
classification success rate (97.7\% vs. 96.2\% and 96.9\%). The MIT 
AI Risk Repository's domain 
taxonomy~\cite{slattery2025airiskrepositorycomprehensive} was 
excluded from the pairwise comparison because categories 
that should have been distinct collapsed into one: 52\% of all 
classifications used a single catch-all category 
(\emph{``AI system safety, failures, and limitations''}), making 
the taxonomy look easy to use even though the categories did not match the risks well.

Qualitative reflections attributed the advantage to workplace 
specificity: our taxonomy names workplace consequences rather than 
AI mechanisms. A software engineer noted: \textit{``Workplace taxonomy went above and beyond and provided the specific result of the 
misinformation, which was lowered efficiency''} (18--29, Male, 
Software Engineer). A family medicine doctor found that it 
\textit{``clearly captured the patient safety consequences of 
prescribing the wrong medication; the GenAI Risk Taxonomy focused 
more on the data/privacy aspect, while the AI Risk Repository's Domain Taxonomy emphasized the technical AI failure rather than the medical risk to 
patients''} (30--39, Female, Family Medicine Doctor). A financial 
analyst highlighted that \textit{``the risk is about 
workflow disruption and productivity loss; the GenAI Risk Taxonomy 
was harder because `Financial \& Business' only indirectly captures 
productivity loss''} (18--29, Female, Financial Analyst). Full results and participant quotes are reported in Supplementary Material M.

\section{Discussion}

We first discuss how our findings extend theory on risk of AI agents (\S\ref{subsec:theoretical}), then translate them into practical guidance for workplace governance (\S\ref{subsec:practical}), and finally examine the study's limitations and directions for future research (\S\ref{subsec:limitations}).

\subsection{Theoretical Implications}\label{subsec:theoretical}

\noindent \textbf{Make agentic decomposition explicit.} Rather than treating an AI system as a black box, our framework models risk as emerging from the properties of three core components (agents, goals, environments) and from the interactions between them and humans. This distinction matters because interaction-driven risks can arise even when no single component is malfunctioning, a pattern our empirical findings confirm. This extends work focused primarily on agent--agent communication~\cite{hammond2025multiagentrisksadvancedai} by showing that goal and environment mediation are equally important risk pathways. Further, the framework decomposes human interaction into five distinct relationship types (agent--human, human--human, human--environment, etc.) rather than treating it as a single category. This enables more precise theorizing about how accountability drifts, how reliance develops, and how distributed agency makes it difficult for workers to contest outcomes that affect them.

\vspace{0.03 in}
\noindent \textbf{Connect framework-level concepts to empirical workplace structure.} By grounding the framework in a 15-category taxonomy built from real job tasks, this paper provides something that prior work largely lacks: a taxonomy that operates at the level of specific workflows rather than broad societal patterns. The severity profile of \emph{Erroneous Actions} by agents, and the dominance of \emph{Capability erosion} risks, together suggest that workplace risks are driven less by model failures than by failures in how agents and humans communicate intent, interpret outputs, and share responsibility. That both leading categories are majority-augmentation reinforces the point: the risk concentrates where a human remains in the loop, not where the agent runs alone. This implies that ``capability'' in agentic systems is not a property of the model alone, but formed by the agent, the human, and the organizational context in which they interact.

\subsection{Practical Implications}\label{subsec:practical}

Beyond showing that our taxonomy is statistically distinct and practically usable (\S\ref{sec:results}, Finding 4), our findings point to three concrete shifts in how organizations should approach AI agent governance.

\vspace{0.03 in}
\noindent \textbf{Focus risk assessment on the human-agent boundary.} The highest-frequency and highest-severity risks in our dataset consistently cluster at the point where humans interpret and act on agent outputs. Governance efforts should therefore not stop at auditing model performance. Assurance processes should also evaluate how uncertainty is communicated, whether workers have clear pathways to question or override agent recommendations, and whether organizational goals and proxy metrics create incentives that the agent system could exploit in harmful ways.

\vspace{0.03 in}
\noindent \textbf{Apply layered controls that match the framework layers.} At technical capability layer, organizations should test how components fail under stress: partial observability simulations, proxy metric sensitivity checks, and multi-agent coordination tests. At human interaction layer, safeguards should address human behavior: role clarity training, contestability channels that allow workers to challenge agent decisions, and structured interventions against over-reliance. At systemic impact layer, monitoring should track slower systemic effects such as deskilling trajectories, shifts in bargaining power, and the adoption dynamics that can normalize unsafe practices across an industry.

\vspace{0.03 in}
\noindent \textbf{Govern automation and augmentation separately.} Automated deployments concentrate risk in institutional outcomes: legal exposure, operational failure, reputational damage. These require hard fail-safes, clear audit trails, and well-defined incident response procedures. Augmented deployments concentrate risk in the individual worker: skill erosion, professional judgment displacement, and workplace stress. These require socio-technical safeguards: protected human override, regular skill assessments, and workflow redesigns that ensure workers retain meaningful agency rather than simply approving agent recommendations. The taxonomy itself can serve as a practical tool here, used as a living risk register that is updated iteratively as agent capabilities and organizational practices evolve.

\subsection{Limitations and Future Work}\label{subsec:limitations}

Although the scenarios were generated by an LLM, we evaluated them with workers and LLM judges and grounded them in real incident data and a live multi-agent red-teaming study. This methodology is well grounded in established risk foresight and validation practice~\cite{perez2022red, ganguli2022red, frohling2026agent, buçinca2023ahafacilitatingaiimpact, 10.1145/3706598.3713979, 10.5555/3716662.3716711, Wang_2024_Farsight}, but future work could test whether the same risk patterns emerge under different generation conditions, prompting techniques, and agentic approaches. The divergence in novelty scores between workers and the LLM judge also deserves attention. Workers found the scenarios moderately novel, offering genuine discovery value; the LLM was far stricter, likely recognizing training data patterns. This raises a question on whether there categories of workplace AI agent risk that are genuinely novel and therefore harder to surface through LLM-driven foresight?

The automation \emph{vs.} augmentation distinction simplifies a continuum. Many real deployments are hybrid: tasks may be partially automated, intermittently supervised, or redistributed across roles in ways that do not fit cleanly into either category. Similarly, O*NET reflects a U.S.-centric occupational framing and may not capture informal work, global labor conditions, or sector-specific practices. Finally, the taxonomy represents a snapshot. Because the scenarios are LLM-generated, the category frequencies and distributional analyses describe only the composition of our risk corpus; they should not be interpreted as estimates of how often these risks occur in real workplaces. These patterns may also reflect the selected job tasks, prompt design, and model used. As agent capabilities evolve, both the distribution of risks and the taxonomy structure will need updating. Future work should triangulate these LLM-generated scenarios with field studies and incident reports, test robustness across models and prompt designs, enrich the occupational grounding with organizational context such as staffing constraints and safety culture, and conduct participatory validation with workers and domain stakeholders to ensure the taxonomy supports fairness-oriented governance in practice.

\section{Conclusion} As AI agents become embedded in everyday work, organizations need practical ways to anticipate the risks they may introduce. This paper contributes: a multi-layer framework that models how risks emerge from interactions between agents, goals, environments, and workers; a corpus of 8,356 risk scenarios grounded in 2,078 real job tasks across 209 roles; and a 15-category taxonomy that organizes those risks under 4 themes. Across our corpus, the main pattern is that workplace risks often emerge from breakdowns at the human--agent boundary rather than from isolated model failures. More scenarios were classified as \emph{Erroneous Actions} of agents than under any other category (30.6\%), and 88.2\% of these scenarios were rated High or Critical. \emph{Capability Erosion} was the second most frequently assigned category (21.3\%). Scenarios in both categories were predominantly associated with augmentation, showing that keeping a worker in the loop does not eliminate risk. These patterns point to the same conclusion: the central challenge is not only building more capable agents, but also designing safer ways for humans to work alongside them.
\newpage
\section*{Acknowledgements}
This work was supported by UK Research and Innovation (grant number EP/S023356/1), in the UKRI Centre for Doctoral Training in Safe and Trusted Artificial Intelligence (www.safeandtrustedai.org).
\bibliography{aaai2026}

\clearpage

\end{document}